\documentclass[twocolumn]{article}
\PassOptionsToPackage{dvipsnames}{xcolor} 
\usepackage{preprint}
\SetBgContents{} 
\usepackage[utf8]{inputenc}
\usepackage[T1]{fontenc}
\usepackage[english]{babel}
\usepackage{csquotes}
\usepackage[final,
            tracking=true,
            kerning=true]{microtype}
\usepackage{xspace}

\usepackage{amssymb}
\usepackage{amsmath}
\usepackage{mathtools}
\usepackage{bbm}

\usepackage{graphicx}
\usepackage{booktabs}
\usepackage{subcaption}
\usepackage{afterpage}

\usepackage{nth}
\usepackage{siunitx}
\usepackage[dvipsnames]{xcolor}

\newcommand*{\titletext}{Recovering Localized Adversarial Attacks}
\title{\titletext}

\usepackage[backend=biber,
            sorting=none,
            maxcitenames=2,
            maxbibnames=3]{biblatex}
\usepackage{varioref}
\usepackage[pdfencoding=auto,
            colorlinks,
            linkcolor=MidnightBlue,
            citecolor=ForestGreen,
            urlcolor=Maroon,
            pdfauthor={Jan Philip Göpfert},
            pdftitle={\titletext}]{hyperref}
\usepackage[all]{hypcap}
\usepackage[noabbrev,
            capitalize,
            nameinlink]{cleveref}

\title{\titletext}

\usepackage[auth-sc]{authblk}

\author[1]{Jan P. Göpfert}
\author[2]{Heiko Wersing}
\author[1]{Barbara Hammer}
\affil[1]{Bielefeld University, Germany}
\affil[2]{Honda Research Institue Europe, Offenbach am Main, Germany}

\newcommand\blfootnote[1]{{%
  \let\thempfn\relax%
  \footnotetext[0]{#1}%
}}

\usepackage{todonotes}
\usepackage{layouts}

\newcommand*{\eg}{e.\,g.\@\xspace}
\newcommand*{\ie}{i.\,e.\@\xspace}
\newcommand*{\cf}{cf.\@\xspace}

\DeclarePairedDelimiterX{\norm}[1]{\lVert}{\rVert}{#1}

\begin{document}
\twocolumn[
\begin{@twocolumnfalse}
\maketitle
\begin{abstract}
Deep convolutional neural networks have achieved great successes over recent
years, particularly in the domain of computer vision. They are fast, convenient,
and -- thanks to mature frameworks -- relatively easy to implement and deploy.
However, their reasoning is hidden inside a black box, in spite of a number of
proposed approaches that try to provide human-understandable explanations for
the predictions of neural networks. It is still a matter of debate which of
these explainers are best suited for which situations, and how to quantitatively
evaluate and compare them~\autocite{mohseni2018survey}. In this contribution, we
focus on the capabilities of explainers for convolutional deep neural networks
in an extreme situation: a setting in which humans and networks
fundamentally disagree. Deep neural networks are susceptible to adversarial
attacks that deliberately modify input samples to mislead a neural network's
classification, without affecting how a human observer interprets the input. Our
goal with this contribution is to evaluate explainers by investigating whether
they can identify adversarially attacked regions of an image. In particular, we
quantitatively and qualitatively investigate the capability of three popular
explainers of classifications -- classic salience, guided backpropagation, and
LIME -- with respect to their ability to identify regions of attack as the
explanatory regions for the (incorrect) prediction in representative examples
from image classification. We find that LIME outperforms the other explainers.
\end{abstract}
\vspace{0.5cm}
\end{@twocolumnfalse}
]
\blfootnote{This work was supported by Honda Research Institute Europe GmbH,
Offenbach am Main, Germany.}
\section{Introduction}
In recent years, deep learning has led to astonishing achievements in several
domains, including gaming, machine translation, speech processing, and computer
vision~\autocite{Schmidhuber2015DeepLI}. The deep neural networks involved act
mostly as black boxes, and as a result they are often met with a certain
wariness, especially in safety-critical environments, matters where fairness is
important, or when rigorous explanations of a decision are legally required. A
number of approaches have been proposed which aim to explain the decisions of
neural networks to human users. They include methods that determine particularly
relevant input regions for a certain decision, methods that locally approximate
complex decisions via human-understandable sparse surrogates, classifier
visualization techniques, or more general methods that supplement automated
decisions by a notion of their domain of expertise, and explicit reject options
whenever their validity is questionable~\autocite{Fischer2016OptimalLR,Ribeiro2016WhySI,Samek2017ExplainableAI,Schulz2014UsingDD}.

Explainers need to address two contradictory goals: they need to preserve the
explained (highly nonlinear) model's behavior as much as possible, but simplify
it such that it becomes accessible to humans in the form of an explanation. In
practice, it is unclear in how far established explainers master this
compromise. One problem is that, given an input and a prediction, it is not
necessarily clear what \emph{a correct explanation} for the prediction should
look like, because the ground truth of which features truly influence the
network's prediction is unknown. It might be tempting to judge explanatory
methods on whether they succeed in identifying features that a human observer
thinks \emph{should} be relevant to the classification, but the existence of
adversarial examples shows that the reasoning of humans and neural networks can
differ dramatically. In this contribution, we exploit the existence of
adversarial examples, using localized adversarial attacks to construct pairs of
inputs and predictions together with ground-truth information about which image
pixels determine the prediction of the network.

Adversarial attacks are an unsolved challenge for deep neural networks -- and
more generally for black-box approaches that aim to classify high-dimensional
data as is present in computer vision. These attacks result in adversarial
examples, which are deliberately generated to fool a classifier. Depending on
the specifics of the attack, it may or may not not be recognizable by humans,
with noticeable artifacts being produced in some
cases~\autocite{HiddenInPlainSight}. In any case, a proper adversarial attack
modifies a given input in such a way that the attacked neural network estimates
a different (wrong) label, while a human user would assign the same label to the
modified input as to the original input.

In this contribution, we use adversarial examples as an extreme setting in which
we can investigate the capabilities of explainers with regards to what makes an
input adversarial. In other words, we want to understand how adversarial attacks
affect explanations of predictions of deep neural networks and make use of them
to produce ground-truth explanations, which allows a quantitative evaluation of
explainers. For this we explain adversarial attacks in
\cref{main:adversarial-attacks}. Then, we take a look at three popular
explainers for neural networks in \cref{main:explainers}, namely:
\begin{itemize}
    \item \emph{classic salience}~\autocite{Selvaraju2017GradCAMVE} maps, which
    are based on gradients propagated through a neural network
    \item \emph{guided backpropagation}~\autocite{Springenberg2014StrivingFS},
    which also takes into account the representations that are implicitly
    learned by neural networks, and
    \item \emph{LIME}~\autocite{Ribeiro2016WhySI}, which locally approximates
    the usually highly nonlinear neural network by a sparse, linear,
    human-understandable surrogate model.
\end{itemize}
In \cref{main:experiments}, we define the setting and evaluate the behavior of
these methods within the field of computer vision: we quantitatively evaluate in
how far methods that explain the decision of deep neural networks can locate
where an adversarial attack has modified an image. We finish with a conclusion
in \cref{main:conclusion}.
\begin{figure*}[htp]
    \centering
    \begin{subfigure}[t]{.26\linewidth}
        \centering
        \includegraphics[width=.9\linewidth]{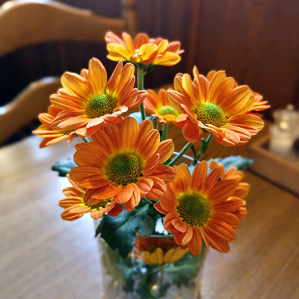}
        \caption{}
    \end{subfigure}
    \begin{subfigure}[t]{.26\linewidth}
        \centering
        \includegraphics[width=.9\linewidth]{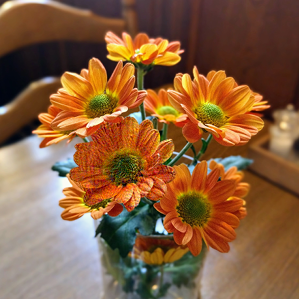}
        \caption{}
    \end{subfigure}
    \begin{subfigure}[t]{.26\linewidth}
        \centering
        \includegraphics[width=.9\linewidth]{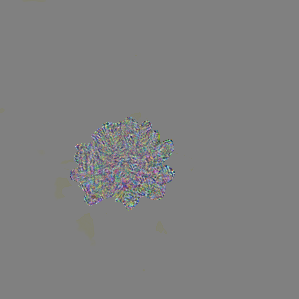}
        \caption{}
    \end{subfigure}
    \caption{An adversarial attack: The original image (a) is classified
             \emph{flower}. After a localized adversarial attack is performed
             the resulting adversarial example (b) is classified \emph{wolf
             spider}. The difference (c) between the two images is seemingly
             random noise. It is confined to one of the blossoms in the image.}
    \label{fig:attack}
\end{figure*}
\section{Adversarial Attacks}\label{main:adversarial-attacks}
Given a classifier \(f\), a sample \(x\) and a label \(y = f(x)\) that \(f\)
assigns to \(x\), the goal of an adversarial attack is to modify \(x\) {just
enough} such that \(f\) assigns a different label \(z\) to the modified sample
\(x'\), where \(z \neq y\) can be \emph{any} other label (in which case the
attack is called \emph{untargeted}) or a \emph{specific} label (in which case
the attack is called \emph{targeted}). In the simplest setting, \(f\) is known
to the attacker in its entirety. Black-box attacks, on the other hand, attack
deep networks without requiring access to \(f\) itself. Instead, they use a
surrogate that is inferred from a representative training set. In this work we
assume that \(f\) is available. Commonly, an untargeted attack on a sample \(x\)
is formalized as the optimization problem
\begin{equation}
    \min \norm{x' - x} \quad \text{such that} \quad f(x') \neq f(x), x' \in C(x),
\end{equation}
where $C$ denotes additional constraints on the adversarial example $x'$, such
as box constraints or sparsity. Early approaches aim for an optimization of this
problem by standard solvers such as LBFGS, while more recent approaches vary the
objective and optimization strategies; software suites available include
\emph{foolbox}~\autocite{1707.04131} and
\emph{cleverhans}~\autocite{1610.00768}.

For our evaluation, we need to efficiently perform targeted adversarial attacks
constrained to varying regions within a number of different input images. This
process yields adversarial examples together with ground truth as to which
region in an example is responsible for its (mis-)classification. We use a
targeted, iterative variant of the
\emph{Fast Gradient Sign Method} (FGSM)~\autocite{Goodfellow2014ExplainingAH},
the \emph{Basic Iterative Method} (BIM)~\autocite{Kurakin2016AdversarialML}.
BIM, just as FGSM, relies on the fact that adversarial attacks can be observed
also for linear mappings in high dimensional input spaces. Based on this
rationale, attacks move an input \(x\) along a linear approximation of the
objective \(J(x,y)\) of the network, adding the change
\(\epsilon \cdot \mathrm{sign}[\nabla_x J(x,y)] \).
\subsection*{Localized attacks}
We want to investigate whether an explanation can identify the attack as the
reason for the prediction of the neural network. For this purpose, we use a modified
version of BIM, a localized attack~\autocite{HiddenInPlainSight}, for which a
quantitative evaluation of the question, whether an explainer identified the
right region, is straightforward: we implement an additional constraint \(C(x)\)
by allowing \(x'\) to deviate from the original input image \(x\) only in a
specified region of \(x\). This enables us to evaluate the explanation of the
resulting adversarial example by measuring the overlap of the pixels (\ie
features) that constitute the explanation with those within the attack region.
\section{Explaining Predictions}\label{main:explainers}
There exist different methods to explain predictions as produced by black-box
mechanisms such as deep networks. Explanations can be either local for the
decision \(f(x)\) for input \(x\), or they can be global for the function \(f\).
They typically focus on either features or prototypes as the basic “language” to
explain the model. Here, we are interested in explaining adversarial examples
generated by changing a limited number of features. Hence, we focus on local
explanations of \(f(x)\) where \(x\) is an adversarial example, and use for
methods that provide a set of features which best explain this decision. Our
quantitative evaluation of the results relies on the overlap of features which
we changed during generation of an adversarial example \(x\), and the set of
features that are used to explain the decision \(f(x)\). We compare three
different local explanation strategies with respect to their ability in
identifying the features where attacks have taken place.
\paragraph*{Classic Salience}
Salience maps were proposed \eg by \textcite{Selvaraju2017GradCAMVE} as a visual
feedback about the most relevant regions of an image for a specific
classification. Essentially, an input feature \(x_i\) is highlighted according
to its relevance for the classification as given by the gradient \(\partial J(x,
y)/\partial x_i\).
\paragraph*{Guided Backpropagation}
One of the reasons for the success of deep convolutional networks is attributed
to their ability to learn higher level feature representations of the object as
represented within the activation of the hidden
layers~\autocite{Bengio2013RepresentationLA}. A plain gradient as used for
salience maps does not focus on these features because it propagates back both
positive and negative contributions of the gradients. Guided
backpropagation~\autocite{Springenberg2014StrivingFS} circumvents this problem
by truncating negative gradients during backpropagation.
\paragraph*{LIME}
\Textcite{Ribeiro2016WhySI} proposed \emph{Local Interpretable Model-agnostic
Explanations} (LIME) as an agnostic method that does not use the specific form
of the classifying function \(f\) that it explains. It tries to approximate the
function \(f\) locally around \(x\) by an interpretable surrogate in the form of
a sparse model in features \(\tilde x_i\) derived from \(x\). For this purpose,
examples are generated around \(x\) by jittering, and labeled according to
\(f\). The resulting training set is used to infer a sparse, explainable, linear
model, which describes \(f\) locally around \(x\). For image classification, the
basic features \(\tilde x_i\) are typically superpixels, which are obtained from
a perceptual grouping of the image pixels.
\section{Experimental Evaluation}\label{main:experiments}
\newcommand*{\numimages}{112}
\newcommand*{\numattacksmax}{10}
\newcommand*{\numadversarialexamples}{608}
\newcommand*{\numexplanations}{12160}
\newcommand*{\numsuperpixels}{20}
We evaluate the information which is provided by these explanations about
adversarial attacks for the popular deep neural network
\emph{Inception~v3}~\autocite{Szegedy2016RethinkingTI} as provided by pytorch's
torchvision package. We are interested in two research questions:
\begin{description}
    \item[R1] Is it possible to uncover substantial information about the
    location of adversarial attacks in an image by means of explainers?
    \item[R2] If the answer is yes, are there substantial differences with
    regards to the effectiveness of different explanation strategies as
    introduced above?
\end{description}
\paragraph{Generating adversarial examples}
\begin{figure*}[tp]
    \centering
    \includegraphics[width=15cm]{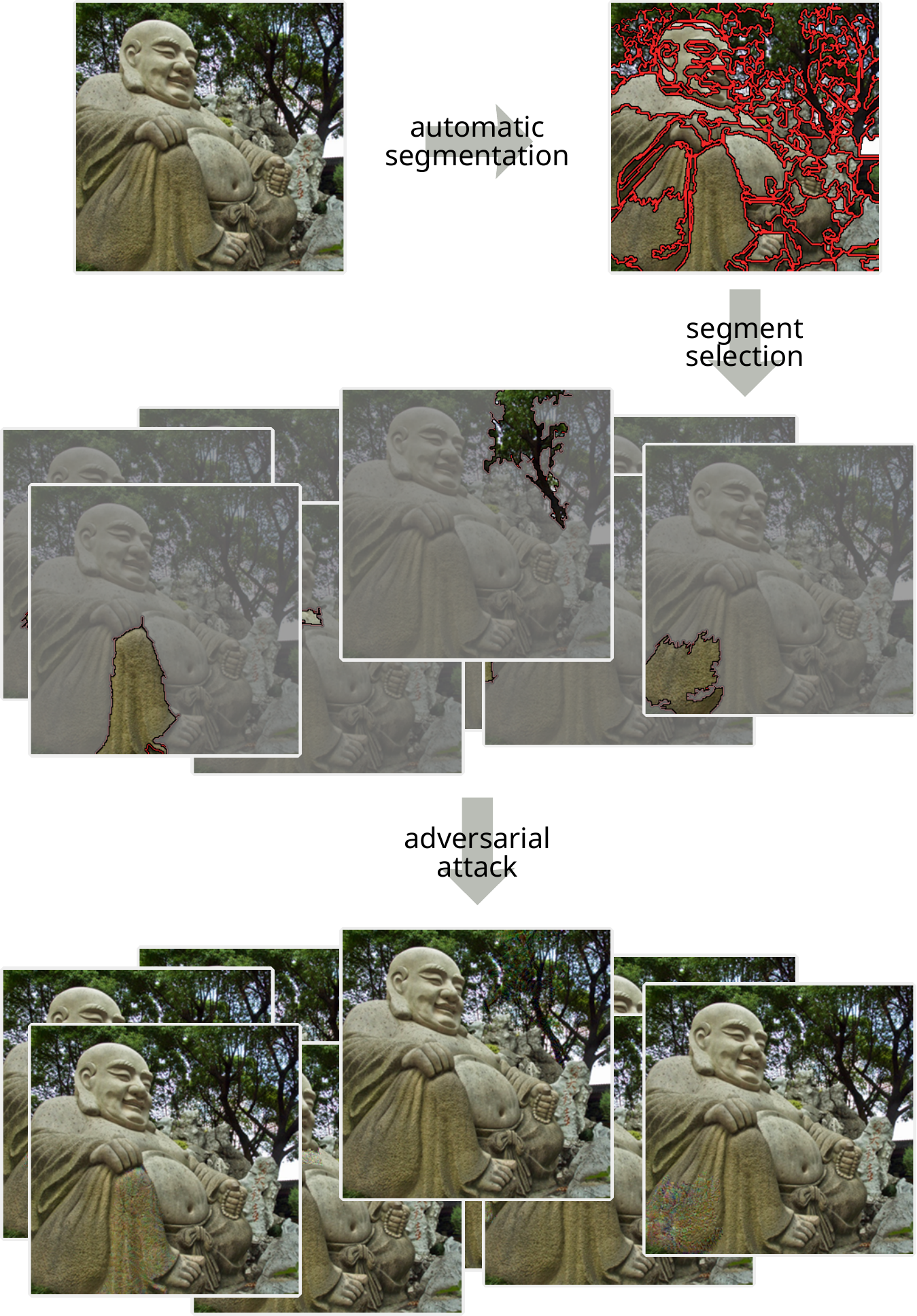}
    \caption{Overview of our process of generating multiple adversarial examples
    via localized adversarial attacks from a single input image. First, the
    image is automatically segmented. Then, the largest segments are chosen and
    individually used to constrain an adversarial attack.}
    \label{fig:example-generation}
\end{figure*}
\begin{figure*}[htp]
    \captionsetup[subfigure]{aboveskip=.3pt}
    \centering
    \begin{subfigure}[t]{.26\linewidth}
        \centering
        \includegraphics[width=.9\linewidth]{media/flowers.png}
        \caption{}
    \end{subfigure}
    \begin{subfigure}[t]{.26\linewidth}
        \centering
        \includegraphics[width=.9\linewidth]{media/flowers-segment1-attacked.png}
        \caption{}
    \end{subfigure}
    \begin{subfigure}[t]{.26\linewidth}
        \centering
        \includegraphics[width=.9\linewidth]{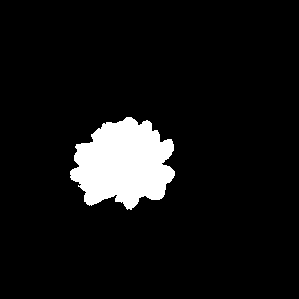}
        \caption{}
    \end{subfigure}
    \begin{subfigure}[t]{.26\linewidth}
        \centering
        \includegraphics[width=.9\linewidth]{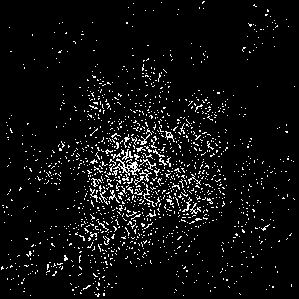}
        \caption{}
    \end{subfigure}
    \begin{subfigure}[t]{.26\linewidth}
        \centering
        \includegraphics[width=.9\linewidth]{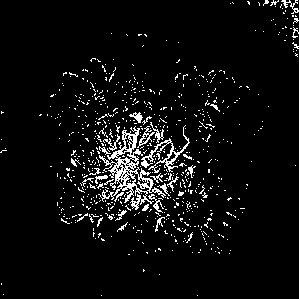}
        \caption{}
    \end{subfigure}
    \begin{subfigure}[t]{.26\linewidth}
        \centering
        \includegraphics[width=.9\linewidth]{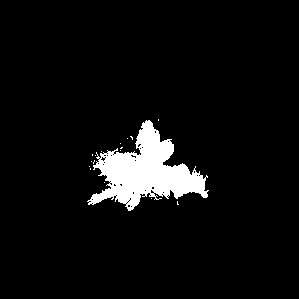}
        \caption{}
    \end{subfigure}
    \caption{Partial explanations for the adversarial example from
    \cref{fig:attack} for classic salience (d), guided backpropagation (e), and
    LIME (f). (c) is the ground truth. White indicates pixels relevant for the
    classification. All explanations contain the same number of white pixels.}
    \label{fig:example-explanation}
\end{figure*}
To guarantee that our test images were not part of the attacked network's
training set, we use crops of \num{\numimages}~images that we took ourselves.
For each attack, we set the constraint \(C(x)\) such that only a relatively
small region within the input image is modified. We obtain those regions by
automatic segmentation using the graph-based algorithm proposed by
\textcite{Felzenszwalb2004EfficientGI} -- during this process, semantics are not
explicitly taken into account, and regions are instead constructed based on
color statistics. When a region contains exactly one object, the attack can
resemble the replacement of said object -- to illustrate this, we manually
segment a small number of input images, \eg the one seen in \cref{fig:attack}.
Out of every original input image we generate up to \numattacksmax~adversarial
examples via BIM restricted to the \numattacksmax~largest regions (\cf
\cref{fig:example-generation}). The same target label \emph{wolf spider} is used
for every attack. If an attack is not successful or ceases to progress after a
certain number of iterations, we discard the attempt. In total, we produce
\num{\numadversarialexamples}~adversarial examples. With this setup we can
guarantee that there exist different regions of the same image, which are
attacked and should be uncovered, i.e.\ finding the location of an attack is a
non-trivial task, which is not already determined by the image itself.
\paragraph{Evaluation of Explanations}
We explain each adversarial example using Classic Salience, Guided
Backpropagation, and LIME. LIME segments the adversarial example into disjoint
superpixels and ranks those by their influence on the prediction. We look at the
\num{\numsuperpixels}~most influential superpixels \(S_1, \dots,
S_{\numsuperpixels}\) and see how well the partial union
\(\mathop{{\bigcup}}_{i=1}^n S_i\) for \(n = 1, \dots, \numsuperpixels\)
recovers the constraint region \(C\). For Classic Salience and Guided
Backpropagation we sort the pixels in the adversarial example by the
\(\ell_1\)~norm of the respective gradients, \ie by their influence on the
prediction. In order to compare the results to those produced by LIME, we look
at the \(\left|\mathop{{\bigcup}}_{i=1}^n S_i\right|\) pixels with the highest
influence (\cf \cref{fig:example-explanation}). In total, we compare
\num{\numexplanations}~explanations for each of the three explainers.

To determine how well such a set of pixels \(P\) recovers the region \(C\) we
calculate the Jaccard Index of the two sets
\begin{equation}
    J(P, C) = \frac{P \cap C}{P \cup C}
\end{equation}
and a likeness
\begin{equation}
    H(P, C) = 1 - \frac{\mathop{\text{Ham}(P, C)}}{N}
\end{equation}
which we base on the Hamming distance between \(P\) and \(C\) interpreted as
binary masks over the entire image with \(N\) pixels in total. Both values are
between zero and one, with one indicating a perfect match.

LIME distinguishes between superpixels that strongly contribute towards a
certain prediction and those that strongly oppose it. We only take into account
the former, and in order to interpret the salience maps accordingly, we discard
negative gradients in the input layer before we calculate the gradients'
magnitudes. 
\paragraph{Results}
\begin{table}[b]
    \centering
    \caption{Mean ranks for all three explainers over all
    \num{\numexplanations}~explanations with respect to the Jaccard index and
    the Hamming-based likeness. \num{1.0}~is best, \num{3.0}~is worst.}
    \begin{tabular}{lrrrr}
        \toprule
                         & \multicolumn{2}{c}{Mean rank} \\
        \cmidrule(l){2-3}
        Explainer        & Jaccard & Hamming \\
        \midrule
        Classic salience & 2.58    & 2.59 \\
        Guided backprop  & 2.06    & 2.03 \\
        LIME             & 1.36    & 1.38 \\
        \bottomrule
    \end{tabular}
    \label{tab:mean-scores}
\end{table}
\begin{figure*}[t]
    \centering
    \includegraphics[width=16cm]{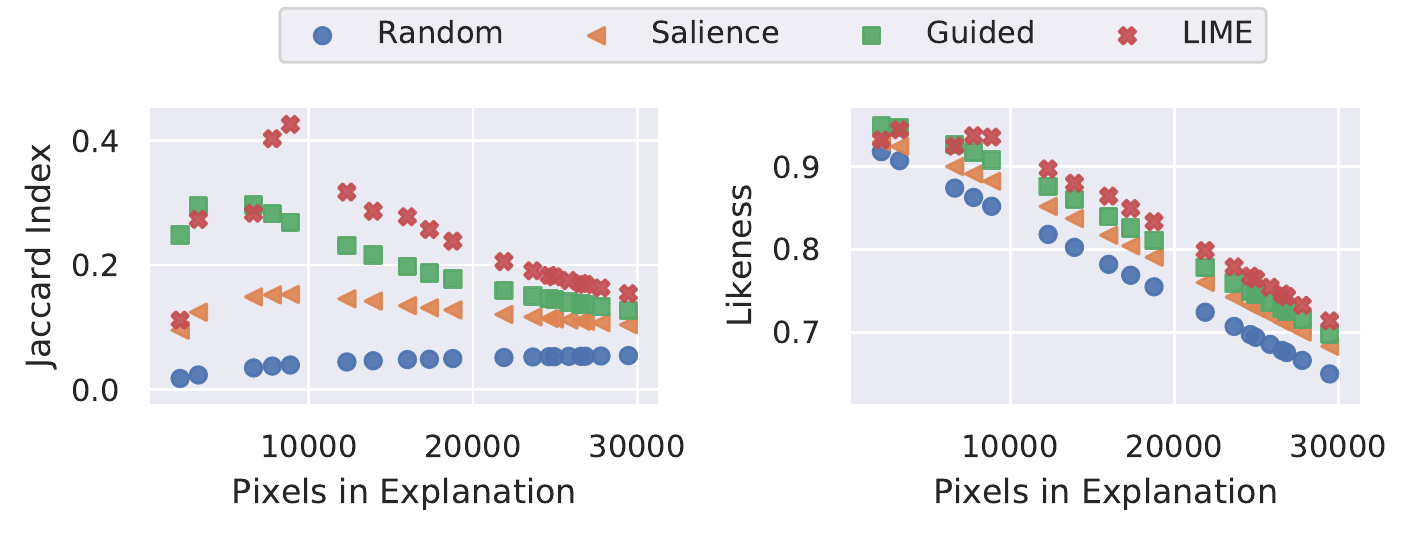}
    \caption{Jaccard index and Hamming-based likeness for different explanation
    sizes for the adversarial example from \cref{fig:attack}.}
    \label{fig:plot}
\end{figure*}
When we compare the explanations provided by LIME with the ground truth, if for
a certain \(n \in \{1, \dots, \numsuperpixels\}\) the partial union
\(\mathop{{\bigcup}}_{i=1}^n S_i\) contains all ground-truth pixels, larger
unions can only perform worse. We see this in \cref{fig:plot}, where the best
Jaccard index is reached early for a relatively small number of pixels in the
explanation. Classic salience and guided backpropagation behave comparably. To
demonstrate that the obtained values are indeed meaningful, we include a random
baseline -- selecting pixels at random yields a very low Jaccard index.

Note that the default segmentation algorithm inside LIME differs from the one we
use to automatically determine regions we attack. Hence, it is almost impossible
for LIME to achieve perfect scores.

To compare all three explainers with regards to the entire
\num{\numadversarialexamples}~adversarial examples, we rank them according to
the Jaccard index and Hamming index for each example from \num{1} (best) to
\num{3} (worst). The mean ranks are listed in \cref{tab:mean-scores}. LIME
outperforms the other two methods, even though those are based on gradients,
just as the adversarial attacks.

All explainers include pixels outside the ground-truth region in their
explanations. This is especially noticable for LIME, where entire contiguous
segments are selected. This is to be expected, because the neural network's
prediction is reached considering the entire input image. Our attacks only
change the prediction from one label to a different one. Pixels outside the
attacked region can still contribute to both.
\section{Conclusion}\label{main:conclusion}
The results in \cref{main:experiments} allow us to answer the initial two
research questions:
\begin{description}
    \item[R1] All three tested explanation techniques detect a substantial part
    of the region where adversarial attacks have taken place which is clearly
    better than random.
    \item[R2] Explanation methods that focus on semantics rather than mere
    gradients, as offered by guided backpropagation and LIME, perform
    distinctly better in the tested settings.
\end{description}
The latter finding is particularly interesting in the sense that saliency is
essentially based on the same information, which also guides adversarial
attacks, namely gradient information. Still, LIME or truncated gradient, both
relying on simplifying assumptions, result in a better recovering of the regions
where attacks have taken place.

We have investigated the behavior of explanatory methods for deep learning when
confronted with adversarial examples. We found that semantics-based approaches
in particular are able to identify a substantial part of regions in which an
attack has taken place, for a representative set of samples.

In general, we desire a better understanding of adversarial attacks, robustness
against them, the certainty of predictions and their explanations, and of how
deep convolutional neural networks divide the input space into class regions.
With this work, we contribute but a small step towards a more comprehensive
grasp of these interlinked concepts. Understanding how labels relate to each
other might allow us to construct ground truth with a clearer distinction
between strongly and weakly relevant pixels, so that pixels outside attacked
regions do not contribute to the prediction as much. Unfortunately, current
state-of-the-art classifiers ignore semantic similarities between classes.

Our findings support the idea that it is possible to recover regions that are --
by design -- the cause for incorrect (adversarial) classifications. In
subsequent work we will investigate whether our findings generalize to
alternative classification methods and whether explanations of adversarial
examples display systematic differences when compared to explanations of proper
(correctly classifiable) samples. Furthermore, we will produce an extension
towards an interactive scenario in which a human user is aided in understanding
principles and pitfalls of automated classification.
\AtNextBibliography{\raggedright}
\printbibliography

@Misc{Goodfellow2014ExplainingAH,
  author     = {Ian J. Goodfellow and Jonathon Shlens and Christian Szegedy},
  title      = {Explaining and Harnessing Adversarial Examples},
  year       = {2014},
  eprint     = {1412.6572},
  eprinttype = {arXiv},
}

@Misc{Springenberg2014StrivingFS,
  author     = {Jost Tobias Springenberg and Alexey Dosovitskiy and Thomas Brox and Martin A. Riedmiller},
  title      = {Striving for Simplicity: The All Convolutional Net},
  year       = {2014},
  eprint     = {1412.6806},
  eprinttype = {arXiv},
}

@InProceedings{Ribeiro2016WhySI,
  author     = {Marco Túlio Ribeiro and Sameer Singh and Carlos Guestrin},
  title      = {“Why Should I Trust You?”: Explaining the Predictions of Any Classifier},
  booktitle  = {Proceedings of the 22nd ACM SIGKDD International Conference on Knowledge Discovery and Data Mining},
  year       = {2016},
  doi        = {10.18653/v1/n16-3020},
  eprint     = {1602.04938},
  eprinttype = {arXiv},
}

@Article{Schmidhuber2015DeepLI,
  author    = {J{\"u}rgen Schmidhuber},
  title     = {Deep Learning in Neural Networks: An Overview},
  journal   = {Neural Networks},
  year      = {2015},
  volume    = {61},
  pages     = {85--117},
  doi       = {10.1016/j.neunet.2014.09.003},
  publisher = {Elsevier},
}

@Misc{Samek2017ExplainableAI,
  author     = {Wojciech Samek and Thomas Wiegand and Klaus-Robert M{\"u}ller},
  title      = {Explainable Artificial Intelligence: Understanding, Visualizing and Interpreting Deep Learning Models},
  year       = {2017},
  eprint     = {1708.08296},
  eprinttype = {arXiv},
}

@Article{Schulz2014UsingDD,
  author  = {Alexander Schulz and Andrej Gisbrecht and Barbara Hammer},
  title   = {Using Discriminative Dimensionality Reduction to Visualize Classifiers},
  journal = {Neural Processing Letters},
  year    = {2014},
  volume  = {42},
  pages   = {27-54},
  doi     = {10.1007/s11063-014-9394-1},
}

@Article{Fischer2016OptimalLR,
  author  = {Lydia Fischer and Barbara Hammer and Heiko Wersing},
  title   = {Optimal local rejection for classifiers},
  journal = {Neurocomputing},
  year    = {2016},
  volume  = {214},
  pages   = {445-457},
  doi     = {10.1016/j.neucom.2016.06.038},
}

@Article{Szegedy2016RethinkingTI,
  author  = {Christian Szegedy and Vincent Vanhoucke and Sergey Ioffe and Jonathon Shlens and Zbigniew Wojna},
  title   = {Rethinking the Inception Architecture for Computer Vision},
  journal = {IEEE Conference on Computer Vision and Pattern Recognition (CVPR)},
  year    = {2016},
  pages   = {2818-2826},
  doi     = {10.1109/cvpr.2016.308},
}

@Article{Selvaraju2017GradCAMVE,
  author  = {Ramprasaath R. Selvaraju and Michael Cogswell and Abhishek Das and Ramakrishna Vedantam and Devi Parikh and Dhruv Batra},
  title   = {Grad-CAM: Visual Explanations from Deep Networks via Gradient-Based Localization},
  journal = {IEEE International Conference on Computer Vision (ICCV)},
  year    = {2017},
  pages   = {618-626},
  doi     = {10.1109/iccv.2017.74},
}

@Misc{1610.00768,
  author     = {Nicolas Papernot and Fartash Faghri and Nicholas Carlini and Ian Goodfellow and Reuben Feinman and Alexey Kurakin and Cihang Xie and Yash Sharma and Tom Brown and Aurko Roy and Alexander Matyasko and Vahid Behzadan and Karen Hambardzumyan and Zhishuai Zhang and Yi-Lin Juang and Zhi Li and Ryan Sheatsley and Abhibhav Garg and Jonathan Uesato and Willi Gierke and Yinpeng Dong and David Berthelot and Paul Hendricks and Jonas Rauber and Rujun Long and Patrick McDaniel},
  title      = {Technical Report on the CleverHans v2.1.0 Adversarial Examples Library},
  year       = {2016},
  eprint     = {1610.00768},
  eprinttype = {arXiv},
}

@Misc{1707.04131,
  author     = {Jonas Rauber and Wieland Brendel and Matthias Bethge},
  title      = {Foolbox: A Python toolbox to benchmark the robustness of machine learning models},
  year       = {2017},
  eprint     = {1707.04131},
  eprinttype = {arXiv},
}

@Article{Bengio2013RepresentationLA,
  author  = {Yoshua Bengio and Aaron C. Courville and Pascal Vincent},
  title   = {Representation Learning: A Review and New Perspectives},
  journal = {IEEE Transactions on Pattern Analysis and Machine Intelligence},
  year    = {2013},
  volume  = {35},
  pages   = {1798-1828},
}

@Article{Felzenszwalb2004EfficientGI,
  author  = {Pedro F. Felzenszwalb and Daniel P. Huttenlocher},
  title   = {Efficient Graph-Based Image Segmentation},
  journal = {International Journal of Computer Vision},
  year    = {2004},
  volume  = {59},
  pages   = {167-181},
  doi     = {10.1023/b:visi.0000022288.19776.77},
}

@Misc{HiddenInPlainSight,
  author     = {Jan Philip Göpfert and Heiko Wersing and Barbara Hammer},
  title      = {Adversarial attacks hidden in plain sight},
  year       = {2019},
  eprint     = {1902.09286},
  eprinttype = {arXiv},
}

@Misc{mohseni2018survey,
  author     = {Sina Mohseni and Niloofar Zarei and Eric D. Ragan},
  title      = {A Survey of Evaluation Methods and Measures for Interpretable Machine Learning},
  year       = {2018},
  eprint     = {1811.11839},
  eprinttype = {arXiv},
}

@Misc{Kurakin2016AdversarialML,
  author     = {Alexey Kurakin and Ian J. Goodfellow and Samy Bengio},
  title      = {Adversarial Machine Learning at Scale},
  year       = {2016},
  eprint     = {1611.01236},
  eprinttype = {arXiv},
}
\end{document}